\documentclass[a4paper,]{article}
\usepackage{lineno,hyperref,vmargin}
\usepackage{amssymb,amsmath,amsthm, array, appendix}
\usepackage{dsfont}
\usepackage[normalem]{ulem}
\usepackage{url}

\newtheorem{theorem}{Theorem}
\newtheorem{lemma}{Lemma}

\newtheorem{corollary}{Corollary}

\newtheorem{remark}{Remark}

\DeclareMathOperator*{\esssup}{ess\,sup}

\modulolinenumbers[5]

\begin{document}

\title{Sample Complexity Result for Multi-category Classifiers of Bounded Variation}
\author{Khadija Musayeva}% <-this % stops a space

\maketitle

% As a general rule, do not put math, special symbols or citations
% in the abstract or keywords.
\begin{abstract}
We control the probability of the uniform deviation between empirical and generalization performances of multi-category classifiers by an empirical $L_1$-norm covering number when these performances are defined on the basis of the truncated hinge loss function. The only assumption made on the functions implemented by multi-category classifiers is that they are of bounded variation ($BV$). For such classifiers, we derive the sample size estimate sufficient for the mentioned performances to be close with high probability. Particularly, we are interested in the dependency of this estimate on the number $C$ of classes. To this end, first, we upper bound the scale-sensitive version of the VC-dimension, the fat-shattering dimension of sets of $BV$ functions defined on $\mathbb{R}^d$ which gives a $O(\epsilon^{-d})$ as the scale $\epsilon$ goes to zero. Secondly, we provide a sharper decomposition result for the fat-shattering dimension in terms of $C$, which for sets of $BV$ functions gives an improvement from $O(C^{\frac{d}{2}+1})$ to $O(C\ln^2(C))$. This improvement then propagates to the sample complexity estimate.
\end{abstract}

% Note that keywords are not normally used for peerreview papers.
%\begin{IEEEkeywords}
%multi-category pattern classification, functions of bounded variation, metric entropy, fat-shattering dimension, sample complexity 
%\end{IEEEkeywords}

% For peer review papers, you can put extra information on the cover
% page as needed:
% \ifCLASSOPTIONpeerreview
% \begin{center} \bfseries EDICS Category: 3-BBND \end{center}
% \fi
%
% For peerreview papers, this IEEEtran command inserts a page break and
% creates the second title. It will be ignored for other modes.

\section{Introduction}
In the VC framework \cite{Vap98}, both for binary and multi-category classification tasks, 
when minimal assumption on the predictive model is made, 
the (optimal) way one controls the uniform convergence of the empirical performance to the generalization one depends on the loss function used based on which these performances are defined. The choice of the loss function leads to an upper bound involving one of capacity measures, the quantity characterizing the rate of the uniform convergence. The seminal work dealt with the standard indicator loss function \cite{VapChe71} leading to bounds involving the VC-dimension as a capacity measure. This was improved in \cite{BarMen02} via the Rademacher complexity since the mentioned capacity measure is upper bounded by the VC-dimension. Classifiers implementing real-valued functions offer a richer setting to the assessment of their classification performance since the latter can be defined based on a family of margin loss functions which can be distinguished into two classes: margin indicator loss function and those that are Lipschitz continuous \cite{KolPan02}. A generalization bound on the basis of the margin indicator loss function was first obtained in \cite{Bar98}, 
%(the function which penalizes all the values of its input below some margin $\gamma >0$ and thus dominates the indicator loss function). His basic was obtained 
and extended to the multi-class case in \cite{Gue07,Gue17}. These bounds are in terms of the empirical $L_\infty$-norm covering number as a capacity measure. For Lipschitz continuous margin loss functions, analogous bounds for general function classes, both in binary and multi-class setting, involve the Rademacher complexity as a capacity measure \cite{KolPan02,KuzMohSye14,Mau16,Gue17,MusLauGue19}. In this paper, for an instance of the above mentioned loss functions, we are interested in the multi-class extension of the uniform Glivenko-Cantelli result, Lemma~10 combined with Lemma~11 of \cite{BarLon95}, a result controlled by an empirical $L_1$-norm covering number. 

The main---and the sole---assumption we make in this paper regarding predictive models is that the functions they implement are of bounded variation. According to Helly's selection theorem the space of bounded variation functions (the $BV$ space) can be compactly embedded in the $L_1$-space \cite{BarPre12,AmbFusPal2000}. In this sense it is relevant to study the uniform convergence over sets of the mentioned space of functions via an $L_1$-norm covering number. Also, the $BV$ space contains other interesting classes of functions such as absolutely continuous, Lipschitz continuous functions as well as a Sobolev space (and mainly has applications in image processing tasks \cite{AubKor06}). The functions implemented by most classifiers (such as support vector machines \cite{CorVap95}, neural networks\cite{AntBar99} and nearest neighbours \cite{KonWei14}) can be said to be of bounded variation, and thus the assumption made is not too restrictive. In the context of learning theory, the case of BV functions on the real line $\mathbb{R}$ has been addressed in \cite{BarKulPos97,AntBar99}, and thus the focus here is on the case when $d>1$. 
%\cite{Meyer01,AubKor06,BerPif10}.

In our extension, we closely follow the combinatorial method of Pollard \cite{Pol84} based on which he derived the rate of the uniform convergence for the classical Glivenko-Cantelli problem via an empirical $L_1$-norm approximation of the set. We then translate this result into a sample complexity estimate (as in \cite{BarLon95}), i.e., for fixed $\epsilon, \delta \in (0,1)$, a minimum sample size $n$ sufficient for the uniform deviation between the empirical and true means to be at most $\epsilon>0$ with probability at least $1-\delta$. The VC theory relies on the independent (and identical) distribution assumption, but it still applies if one replaces the independence assumption with the asymptotic independence one, the condition satisfied by the so-called mixing processes\cite{Bra86,Yu94}. Following \cite{Meir2000}, we also extend the mentioned results to such processes, where instead of sample size we now deal with the number of independent blocks (or efficient sample size). 

No matter the setting mentioned above, the main focus of the present work is on elaborating the dependency of a sample complexity estimate on the number $C$ of classes. The first step towards this goal is the estimation of the metric entropy of sets of the BV space. Such a bound exists in the $L_1$-norm \cite{Dutta18}, and can be shown to hold with respect to the $L_1(\mu)$-norm for all probability measures $\mu$ with Lebesgue densities. 
%The bound holding with respect to the empirical version of these measures, on the other hand, can be obtained for large values of $n$ which is ineffective. 
Instead, we upper bound the fat-shattering dimension of sets of the $BV$ space which then can be substituted in metric entropy bounds for general function classes \cite{AloBenCesHau97,BarKulPos97,MenVer03,MusLauGue19} which hold for any probability measure on the domain $\mathbb{R}^d$ of the mentioned functions. We obtain a bound scaling as a $O(\epsilon^{-d})$ as $\epsilon \rightarrow 0$. To make explicit the dependency on the number of classes we appeal to a particular bound, the decomposition of capacity measure which allows one to upper bound a capacity measure of a composite class by a set of that of basic classes. Since we are dealing with two capacity measures, the covering number (or the metric entropy) and the fat-shattering dimension, and since they are related to each other via a combinatorial bound (or the metric entropy bound), one can perform the decomposition at the level of either of them.
Decomposition results exist for covering numbers and the fat-shattering dimension \cite{Gue17,Duan12}. The main contribution of this paper is a new efficient decomposition result for the fat-shattering dimension which scales with $C$ as a $O(C \ln^2(C))$. This is an improvement over that in \cite{Duan12} applied in the multi-class setting where the dependency on $C$ worsens with the growth rate of the fat-shattering dimension of basic classes, and particularly, for sets of the $BV$ space it is a $O(C^{\frac{d}{2}+1})$. This decomposition leads to a new, {\em dimension-free}, i.e., not depending on the sample size $n$ metric entropy bound with a sharper dependency on $C$ compared to Corollary~1 in \cite{MusLauGue19}. The application of this result gives a sample complexity estimate with a $O(C \ln^2(C))$ dependency improving upon a $O(C^{d})$ obtained based on the decomposition of the $L_p(\mu)$-norm metric entropies with $p \in \{1,2\}$, and $O(C \ln^{d+2} C)$ with $2 \leqslant p < \infty$. For that in $p=\infty$, our bound gives a comparable result.

The rest of the paper is organized as follows. In Section~\ref{sec:preliminaries} we introduce the theoretical background. Section~\ref{sec:uniform-convergence-L1} is dedicated to upper bounding the probability of the uniform deviation between empirical and generalization performances of the classifiers of interest by an empirical $L_1$-norm. Section~\ref{sec:bv-space-metric-entropy} discusses metric entropy bounds for sets of the $BV$ space, and for such sets derives a new upper bound on the fat-shattering dimension. In Section~\ref{sec:sample-complexity} we introduce an efficient decomposition of the fat-shattering dimension based on which we derive a sample complexity estimate and compare it with the estimate obtained via the decomposition of the empirical $L_\infty$-norm metric entropy. Conclusions and ongoing research are given in Section~\ref{sec:conclusions}. Finally, the case of mixing processes is addressed in Appendix.

\section{Preliminaries} \label{sec:preliminaries}

We consider $C$-category pattern classification problems with finite $C>2$.
Each object is represented by its description $x \in \mathcal{X}$ and the categories $y$ belong to the set $\mathcal{Y}=\{ 1, \dots, C \}$. The goal is to assign each $x$ to one of the categories. Let $Z = \left ( X,Y \right )$ be a random pair with values in $\mathcal{Z} = \mathcal{X} \times \mathcal{Y}$, distributed according to an unknown probability measure $P$. The only information about $P$ is given by an $n$-sample $\mathbf{Z}_n = \left( Z_i  \right)_{1 \leqslant i \leqslant n} =\left( \left ( X_i, Y_ i \right) \right)_{1 \leqslant i \leqslant n}$ made up of $n$ independent copies of $Z$.

The classifiers considered in the present manuscript are defined based on classes $\mathcal{G} = \prod_{k=1}^C \mathcal{G}_k$ of
functions mapping $\mathcal{X}$ to a hypercube in $\mathbb{R}^C$ and a decision rule which for each $g = \left ( g_{k} \right )_{1 \leqslant k \leqslant C}
\in \mathcal{G}$ and for each $x \in \mathcal{X}$, returns either the index of the component function whose value is the highest
%, $\text{dr}_{g}(x)=\argmax_{y \in \mathcal{Y}}g_y(x)$, 
or a dummy category $*$ in case of ties. The classification performance of such classifiers can be assessed based on functions computing the difference between two component functions, 
$$\forall (x,y) \in \mathcal{Z}, \quad f_{g}(x,y)= \frac{1}{2}(g_y \left(x\right) -\max_{k \neq y} g_k \left( x \right)),$$
and a margin loss function which penalizes all values below some margin $\gamma>0$. The performance of most well-known classifiers such as neural networks \cite{AntBar99}, support vector machines \cite{CorVap95}, nearest neighbours \cite{KonWei14} and boosting method \cite{Sch98} can be studied in this margin framework. 

The margin loss function used here is the (parametrized) truncated hinge loss defined as
\begin{align*}
\phi_{\gamma}(t)= \begin{cases} 
                    1, \quad &t \leqslant 0, \\
                    1-\displaystyle{\frac{t}{\gamma}}, \quad &t \in (0, \gamma],\\
                    0, \quad &t>\gamma,
                  \end{cases}
\end{align*}
%$$\phi_{\gamma}(t)=\mathds{1}_{\{t\leqslant 0\}}+\left(1-\frac{t}{\gamma}\right)\mathds{1}_{\{t \in (0, \gamma]\}},  \quad \gamma \in (0, 1], t \in \mathbb{R}, $$
where $\gamma \in (0, 1]$ and $t \in \mathbb{R}$, which is $\displaystyle{\frac{1}{\gamma}}$-Lipschitz continuous.
This loss function is ``insensitive'' to the values of its argument strictly below zero and above $\gamma$ in the sense that, if instead of functions $f_g$, we  use their truncated versions
$$
f_{g,\gamma} \left (x, y\right ) = \max\left(0, \min\left(\gamma, \frac{1}{2}(g_y \left(x\right) -\max_{k \neq y} g_k \left( x \right)) \right) \right),
$$
where $(x,y)\in \mathcal{Z}$, it holds $$\phi_{\gamma}(f_g(z))=\phi_{\gamma}(f_{g,\gamma}(z)), \quad \forall z \in \mathcal{Z}.$$ We denote $\mathcal{F}_{\mathcal{G}} = \left\{f_{g} : g \in \mathcal{G} \right\}$, and $\mathcal{F}_{\mathcal{G},\gamma} = \left\{f_{g,\gamma} : g\in \mathcal{G} \right\}$ for fixed $\gamma \in (0,1]$. This kind of transitioning from $f_g$ to $f_{g,\gamma}$ results in tighter upper bounds in terms of the co-domain now shrinked to $[0, \gamma]$.

With these definitions at hand, we now can define the margin risk of every $g \in \mathcal{G}$ as 
$$
L_{\gamma}\left(g\right) = \mathbb{E}\left[\phi_{\gamma}\left(f_{g,\gamma}\left(Z\right)\right) \right],
$$ 
and its empirical margin risk as 
$$L_{\gamma,n}\left(g\right)= \frac{1}{n} \sum_{i=1}^n \phi_{\gamma}\left(f_{g,\gamma} \left( Z_i \right)\right).$$ 
We are interested in the rate of convergence of the empirical margin risk to the margin risk uniformly over $\mathcal{G}$. The rate of this convergence is controlled by the quantity called {\em capacity} of a classifier: the higher the capacity, the slower the convergence. In this paper, we deal with several well-known capacity measures: covering/packing numbers \cite{KolTih61}, the fat-shattering dimension \cite{KeaSch94} and the (empirical) Rademacher complexity. They are defined below.

We denote by $\mathcal{F}$ a class of real-valued functions on some metric space $\left(\mathcal{T},\rho\right)$. Denote by $\bar{\mathcal{F}} \subseteq \mathcal{F}$ a (proper) $\epsilon$-net of $\mathcal{F}$ with respect to the metric $\rho$:
$$\forall f\in\mathcal{F}, \exists \bar{f} \in \bar{\mathcal{F}}, \quad \rho(f, \bar{f}) < \epsilon.$$  The covering number of $\mathcal{F}$, $\mathcal{N} \left ( \epsilon, \mathcal{F}, \rho \right )$, is the smallest cardinality of $\epsilon$-nets of $\mathcal{F}$. Related to the covering number is the notion of packing number. A subset $\mathcal{F}' \subseteq \mathcal{F}$ is {\em $\epsilon$-separated} with respect to the metric $\rho$ if for any two distinct elements 
$f_1,f_2 \in \mathcal{F}^{'}$, $\rho(f_1,f_2) \geqslant \epsilon$. The $\epsilon$-packing number $\mathcal{M}\left(\epsilon, \mathcal{F}^{\prime}, \rho \right)$ of $\mathcal{F}$ is the maximal cardinality of its $\epsilon$-separated subsets. The (pseudo-)metric of interest in this work is the empirical one: for any $f,f' \in\mathcal{F}$ and $\mathbf{t}_n =(t_i)_{1\leqslant i\leqslant n}\in \mathcal{T}^n$, define
$d_{p, \mathbf{t}_n}$ as
$$
d_{p, \mathbf{t}_n} ( f, f')
= \left ( \frac{1}{n}\sum_{i=1}^n
\left| f (t_i)-f'(t_i)\right|^p\right)^{\frac{1}{p}}, 
\forall p \in [1, +\infty)$$
and 
$
d_{\infty, \mathbf{t}_n} ( f, f')
= \max_{1 \leqslant i \leqslant n} \left |f (t_i)-f'(t_i)\right |.
$
Note that, since $d_{p, \mathbf{t}_n} ( f, f') \leqslant d_{q, \mathbf{t}_n} ( f, f')$ for any $p \leqslant q$, there holds 
\begin{align}
\mathcal{N} \left ( \epsilon, \mathcal{F}, d_{p,\mathbf{t}_n} \right ) \leqslant \mathcal{N} \left ( \epsilon, \mathcal{F}, d_{q,\mathbf{t}_n} \right ). \label{eq:norm-ordering}
\end{align}
We denote $\mathcal{N}_p \left ( \epsilon, \mathcal{F}, n\right )=\sup_{\mathbf{t}_n \in \mathcal{T}^n} \mathcal{N} \left ( \epsilon, \mathcal{F}, d_{p,\mathbf{t}_n} \right )$, and similarly for packing numbers. 
The logarithm of covering number is called metric entropy.

For $\epsilon>0$, a subset $\left \{ t_i: 1 \leqslant i \leqslant n \right \}$
of $\mathcal{T}$
is said to be {\em ${\epsilon}$-shattered} by $\mathcal{F}$ if
there is a witness
$s : \mathcal{T} \rightarrow \mathbb{R}$
such that for any
$( b_i)_{1 \leqslant i \leqslant n}
\in \left \{ -1, 1 \right \}^n$, there is a function
$f \in \mathcal{F}$ satisfying: 
$$
\forall i \in \{1,\dots, n\}, \;\;
b_i \left( f (t_i) - s(t_i)\right) \geqslant \epsilon.
$$
%The vector $\mathbf{b}_n$ is called a {\em witness} to
%the ${\gamma}$-shattering.
The {\em fat-shattering dimension} of $\mathcal{F}$ at scale $\epsilon$, $d_{\mathcal{F}}\left(\epsilon\right)$,
is the maximal cardinality of a subset of $\mathcal{T}$
${\epsilon}$-shattered by $\mathcal{F}$, if such a maximum exists, otherwise $\mathcal{F}$ is said to have infinite fat-shattering dimension at scale $\epsilon$. 

Let $(\sigma_i)_{1 \leqslant i \leqslant n}$ be a sequence of independent random variables taking values in $\{-1,1\}$ with equal probability. The empirical Rademacher complexity of $\mathcal{F}$ given $(t_i)_{1 \leqslant i \leqslant n} \in \mathcal{T}^n$ is defined as
\begin{align*}
\hat{R}_n\left(\mathcal{F}\right)=\mathbb{E}\left[\sup_{f \in \mathcal{F}}\frac{1}{n}\sum_{i=1}^n\sigma_i f\left(t_i\right) \right],
\end{align*}
where $\mathbb{E}$ denotes the expected value.

In this work, we make the regularity assumption on the classes of component functions that they are of bounded variation. To introduce the space of functions of bounded variation, we shall first give the definitions of {\em Lebesgue}, and {\em Sobolev} spaces.
For a broader account on the mentioned spaces the reader may consult\cite{Adams03,Gilbarg15,Zie12,Attouch14}.

Let $\mathcal{F}$ denote the set of all real-valued measurable functions on a measure space $\left(\mathcal{T}, \mathcal{A}, \mu\right)$. For all $p \in \mathbb{N} \setminus \{0\}$, $L_{p}\left(\mathcal{T}, \mathcal{A}, \mu\right)$ is the Lebesgue space of (equivalence classes of) $p$-summable functions $f \in \mathcal{F}$:
\begin{align*}
%&L_p\left(\mathcal{T}, \mathcal{A}, \mu\right)=\\&\left\{f \in \mathcal{F} \; : \; 
\left\Vert f \right\Vert_{L_{p}(\mu)}=\left(\int_{\mathcal{T}} \left| f(t) \right |^p d\mu(t) \right)^{\frac{1}{p}}< \infty 
%\right\}
,
\end{align*}
and 
\begin{align*}
L_{\infty}\left(\mathcal{T}, \mathcal{A}, \mu\right)\!=\!\left\{ f \in \mathcal{F} : \left\Vert f \right\Vert_{L_{\infty}(\mu)}\!=\!\esssup_{t \in \mathcal{T}} \left|f(t)\right| < \infty  \right\}\!,
\end{align*}
where $\esssup_{t \in \mathcal{T}} \left|f(t)\right|\!=\!\inf_{a \geqslant 0} \mu\{t \in \mathcal{T}\!:\!|f(t)|>a \}\!=\!0$. %When no confusion arises, 
We abbreviate $L_{p}\left(\mathcal{T}\right)=L_{p}\left(\mathcal{T}, \mathcal{A}, \mu\right)$, and denote the metric induced from the norm $\lVert \cdot \rVert_{L_{p}(\mu)}$ by $d_{L_p(\mu)}$.
% When $\mu$ is an empirical measure $\frac{1}{n} \sum_{i=1}^n \delta_{t_i}$ supported on $(t_i)_{1 \leqslant i \leqslant n} \in \mathcal{T}^n$, we will drop the measure notation in the definition of the norm.
In the rest of the section we assume $\mu$ to be the Lebesgue measure, which in the case of Euclidean spaces coincides with the definitions of length, area and volume. In such a case, we will drop the measure from the notation of the norm. 

Let $\mathcal{T}$ be an open subset of $\mathbb{R}^d$. Let $k \in \mathbb{N}$ and let $\alpha=(\alpha_i)_{1 \leqslant i \leqslant d} \in \mathbb{N}^d$ be a multi-index with $|\alpha| =\sum_{i=1}^d\alpha_i\leqslant k$. For any $t=(t_i)_{1 \leqslant i \leqslant d} \in \mathcal{T}$, the partial derivatives are denoted by $D_i=\frac{\partial}{\partial t_i}$ and the higher order partial derivatives by
$$D^{\alpha} = \frac{\partial^{|\alpha|}}{\partial t^{\alpha_1}_1 \dots \partial t^{\alpha_d}_d}.$$ The gradient of a real-valued function $f$ on $\mathbb{R}^d$ is denoted by $Df=\left(D_i f\right)_{1 \leqslant i \leqslant d}$. Denote by $C^m(\mathcal{T})$ $m$-times continuously differentiable real-valued functions, and abbreviate $C(\mathcal{T})=C^0(\mathcal{T})$.
Let $\left(C^m_{c}(\mathcal{T}),\mathbb{R}^d\right)$ be the set of $m$-times continuously differentiable functions from $\mathcal{T}$ to $\mathbb{R}^d$ with compact support contained in $\mathcal{T}$.
%, and for $m=0$, we abbreviate $\left(C_{c}(\mathcal{T}),\mathbb{R}^d\right)=\left(C^0_{c}(\mathcal{T}),\mathbb{R}^d\right)$. 
For a given $\alpha$ and for a given $f \in L_1(\mathcal{T})$, $D^{\alpha}f=f_w \in L_1(\mathcal{T})$ is called the $\alpha$-th weak derivative of $f$, if for all $\phi \in \left(C^1_{c}(\mathcal{T}),\mathbb{R}\right)$,
$$
\quad \int_{\mathcal{T}} f(t) D^{\alpha}\phi(t)dt=(-1)^{|\alpha|} \int_{\mathcal{T}} \phi(t) f_w(t) dt.
$$
% where $$D^{\alpha} \phi(x) = \frac{\partial^{|\alpha|} \phi(x)}{\partial x^{\alpha_1}_1 \dots \partial x^{\alpha_d}_d}, \quad x=(x_i)_{1 \leqslant i \leqslant d} \in \mathcal{\mathcal{X}}.$$

Let $p \in [1,\infty]$. Denote by $\left(W^{k,p}\left(\mathcal{T}\right), \lVert \cdot \rVert_{W^{k,p}}\right)$ the space of functions $f \in L_p(\mathcal{T})$ with $D^{\alpha} f$  (in the sense of weak derivative) in $L_p(\mathcal{T})$
%$$D^{\alpha} f(x) = \frac{\partial^{|\alpha|} f(x)}{\partial x^{\alpha_1}_1 \dots \partial x^{\alpha_d}_d}, \quad x=(x_i)_{1 \leqslant i \leqslant d} \in \mathcal{\mathcal{X}}$$
and with the norm $\lVert f \rVert_{W^{k,p}}$ defined as
$$
\lVert f \rVert_{W^{k,p}} = \int_{\mathcal{T}} \sum_{0 \leqslant |\alpha| \leqslant k} |D^{\alpha} f(t)|^p dt=\sum_{0 \leqslant |\alpha| \leqslant k} \lVert D^{\alpha} f \rVert^p_{L_{p}(\mathcal{T})}
$$
for $p \in [1,\infty)$
and 
$$
\lVert f \rVert_{W^{k,\infty}} = \max_{0 \leqslant |\alpha| \leqslant k}\lVert D^{\alpha} f \rVert_{L_{\infty}\left( \mathcal{T}\right)}.
$$ $W^{k,p}$ is called the Sobolev space of integer order. Now we are ready to give the definition of the $BV$ space.

A function $f$ on $\mathcal{T}$ is said to be of bounded variation if and only if it is in $L_1(\mathcal{T})$,
%$$%\int_{\mathcal{X}} f(x) \mbox{div}\phi(x)dx=- \int_{\mathcal{X}} \phi(x), D_if(x) dx,$$
and $Df$ is a finite (vector) Radon measure (i.e., for any $i$ and for every Borel set $B$, 
$$D_i f(B)=\sup\limits_{\substack{K \subset B \\ K \mbox{ is compact}}} D_i f(K),$$ see page 256 in \cite{Fre2000}), such that for all $\phi \in \left(C^1_{c}(\mathcal{T}),\mathbb{R}\right)$ and for all $i \in \{1, \dots, d\}$
$$
\quad \int_{\mathcal{T}} f(t) D_i\phi(t)dt=-\int_{\mathcal{T}} \phi(t) D_i f(t) dt.
$$
Let $\phi=(\phi_i)^{d}_{i=1}$ with $\phi_i \in C^1_{c}(\mathcal{T})$, then $\phi \in \left(C^1_{c}(\mathcal{T}),\mathbb{R}^d\right)$. Let $\displaystyle{\mbox{div}\phi(t)=\sum_{i=1}^d\frac{\partial \phi_i(t)}{\partial t_i}}$ and $\lVert \phi \rVert_{\infty}=\left(\sum_{i=1}^d\sup_{t \in \mathcal{T}} |\phi_i(t)|^2\right)^{\frac{1}{2}}$. The total variation $\left|Df \right|(\mathcal{T})$ of $Df$ is defined as
%the weak gradient of $f$ such that
%$$
%\int_{\mathcal{X}} f(x) \mbox{div}\phi(x)dx=- \int_{\mathcal{X}} \langle \phi(x), Df(x)%\rangle dx.
%$$
\begin{align*}
\left|Df \right|\left(\mathcal{T}\right)=\sup \left\{ \int_{\mathcal{T}} f(t) \mbox{div}\phi(t)dt : \phi \in \left(C^1_{c}(\mathcal{T}),\mathbb{R}^d\right), \lVert \phi \rVert_\infty \leqslant 1  \right\}
\end{align*}
or equivalently as
$$
\left|Df \right|\left(\mathcal{T}\right)=\sup \left\{\sum_{i=1}^d \int_{\mathcal{T}} \phi_i(t)  D_i f(t) dt : \phi \in \left(C_{c}(\mathcal{T}),\mathbb{R}^d\right), \lVert \phi \rVert_\infty \leqslant 1  \right\}\!.
$$
The set of all functions of bounded variation on $\mathcal{T}$ is denoted by $BV(\mathcal{T})$. By $BV(\mathcal{T}, B)$ we will denote the set of all bounded variation functions from $\mathcal{T}$ to $B$. By definition, for any $f \in BV(\mathcal{T})$, $\left|Df \right|\left(\mathcal{T}\right)< \infty$. %The $BV$ space contains absolutely continuous, and Lipschitz continuous functions, and 
It also holds that $W^{1,1}(\mathcal{T}) \subset BV(\mathcal{T})$. 

For the rest of the paper, for all $k \in \mathcal{Y}$, we let $\mathcal{G}_k$ be a class $\mathcal{G}_0 \subseteq BV([0,A]^d, [0,M])$ with $A,M \geqslant 1$ and of total variation $V$. Clearly, $M \leqslant V$. Also, to avoid measurability problems, we assume that all real-valued functions in this paper satisfy image-admissible Suslin condition \cite{Dud84}.

\section{Uniform Convergence via Empirical $L_1$-norm Covering Number}\label{sec:uniform-convergence-L1} 

Following the combinatorial method of Pollard \cite{Pol84}, we extend Lemma~10 combined with Lemma~11 of Bartlett and Long \cite{BarLon95} to the multi-category setting. This gives a result where the scale of the covering number involves the margin parameter $\gamma$ due to the use of a margin loss function. Unlike Pollard, and as in \cite{BarLon95}, we do not eliminate the additional sample introduced in the proof, as a result the exponential factor is reduced at the cost of making the covering number depend on $2n$ points. However, the latter has no impact when using dimension-free combinatorial bounds. The extension of this result to mixing processes is given in \ref{app:beta-mixing}.

\begin{theorem}\label{theo:uniform-convergence-L1}
Fix $\epsilon \in (0,1)$ and $\gamma \in (0, 1]$. Then for any $n>\frac{2}{\epsilon^2}$, there holds
\begin{align}
P^{n}\left(\sup_{g \in \mathcal{G}} \left( L_{\gamma}\left(g\right)- L_{\gamma, n}\left(g\right) \right)  > \epsilon \right) \leqslant 2\mathcal{N}_{1}\left(\frac{\epsilon\gamma}{8},\mathcal{F}_{\mathcal{G},\gamma}, 2n\right)\exp\left(-\frac{n\epsilon^2}{32}\right).\label{eq:uniform-convergence-L1}
\end{align}
\end{theorem}
\begin{proof} [Proof sketch]
The proof is based on the following steps.

1)% Since the truncated hinge loss function dominates the standard indicator loss function, for any $\gamma \in (0,1]$ and for any $g \in  \mathcal{G}$, $L(g) \leqslant L_{\gamma}(g)$, and thus
% $$P^{n}\left(\sup_{g \in \mathcal{G}} \left( L(g)- L_{\gamma, n}(g) \right)  > \epsilon \right) \leqslant P^{n}\left(\sup_{g \in \mathcal{G}} \left( L_{\gamma}(g)- L_{\gamma, n}(g) \right)  > \epsilon \right).$$ 
Apply the symmetrization technique of Vapnik and Chervonenkis \cite{VapChe71} to the the left-hand side of \eqref{eq:uniform-convergence-L1} to bound it by
\begin{align}
2P^{2n}\left\{\sup_{g \in \mathcal{G}} \left(\frac{1}{n}\sum_{i=1}^n \left(\phi_{\gamma}\left( f_{g,\gamma}\left(Z^{\prime}_{i}\right)\right)\!-\! \phi_{\gamma}\left(f_{g,\gamma}\left(Z_i\right)\right)\right)\right) \geqslant \frac{\epsilon}{2} \right\} \nonumber.
%\\ &=
%2\int_{\mathcal{Z}^{2n}}\!\mathds{1}\left\{\sup_{g \in \mathcal{G}} \left(\frac{1}{n}\sum_{i=1}^n \phi_{\gamma}\left( f_{g,\gamma}\left(z^{\prime}_{i}\right)\right)\!-\!\frac{1}{n}\sum_{i=1}^n \phi_{\gamma}\left(f_{g,\gamma}\left(z_i\right)\right)\right) \geqslant \frac{\epsilon}{2} \right\}dP^{2n}(\mathbf{z}_{2n}), \label{eq:first-symmetrization}
\end{align}
where $(Z'_i)_{1 \leqslant i \leqslant n}$ is a sequence of $n$ independent copies of $Z$ also called a ``ghost" sample. 
 
2) Approximate $\mathcal{F}_{\mathcal{G},\gamma}$ by its finite cover with respect to the empirical $L_1$-norm. Let $\bar{\mathcal{G}}$ be a subset of $\mathcal{G}$ so that $\mathcal{F}_{\bar{\mathcal{G}},\gamma}$ is an $\displaystyle{\frac{\epsilon\gamma}{8}}$-net of $\mathcal{F}_{\mathcal{G},\gamma}$ of minimal cardinality $\mathcal{N}\left(\frac{\epsilon\gamma}{8}, \mathcal{F}_{\mathcal{G},\gamma}, d_{1, \mathbf{z}_{2n}}\right)$, i.e., for all $g \in \mathcal{G}$, there exists $\bar{g} \in \bar{\mathcal{G}}$
\begin{align*}
&\frac{1}{2n} \sum_{i=1}^{n} \left(|f_{\bar{g},\gamma}\left(z_i\right)-f_{g,\gamma}\left(z_i\right) |+\left|f_{\bar{g},\gamma}\left(z^{\prime}_{i}\right)- f_{g,\gamma}\left(z^{\prime}_{i}\right)\right|\right) < \frac{\epsilon \gamma}{8}\!.
\end{align*}
%To keep the notation simple, let $\phi^{\prime}_{g}=\phi_{\gamma}(f_{g,\gamma})$. 
At this step we make use of the $\displaystyle{\frac{1}{\gamma}}$-Lipschitz property of $\phi_{\gamma}$:
\begin{align*}
&\frac{1}{2n} \sum_{i=1}^{n}| \phi_{\gamma}(f_{\bar{g},\gamma}\left(z_i\right))-\phi_{\gamma}(f_{g,\gamma}\left(z_i\right)) |+\frac{1}{2n} \sum_{i=1}^{n} \left|\phi_{\gamma}(f_{\bar{g},\gamma}\left(z^{\prime}_{i}\right))-\phi_{\gamma}(f_{g,\gamma}\left(z^{\prime}_{i}\right))\right| \\ &\!\leqslant\!\frac{1}{2n\gamma} \sum_{i=1}^{n} \left(|f_{\bar g,\gamma}\left(z_i\right)-f_{g,\gamma}\left(z_i\right) |+\left|f_{\bar g,\gamma}\left(z^{\prime}_{i}\right)- f_{g,\gamma}\left(z^{\prime}_{i}\right)\right|\right) < \frac{\epsilon}{8}.
\end{align*}
On the other hand,
\begin{align*}
\frac{1}{n} \sum_{i=1}^{n}\phi_\gamma(f_{\bar{g},\gamma}(z_i))-\phi_\gamma(f_{\bar{g},\gamma}(z^\prime_{i})) +\frac{1}{n} \sum_{i=1}^{n} \phi_\gamma(f_{g,\gamma}\left(z^\prime_{i}\right))-\phi_\gamma(f_{g,\gamma}(z_i))< \frac{\epsilon }{4}.
\end{align*}
It follows that 
\begin{align*}
\frac{1}{n} \sum_{i=1}^{n}\! \left(\phi_\gamma(f_{g,\gamma}\left(z^\prime_{i}\right))\!-\!\phi_\gamma(f_{g,\gamma}(z_i)) \right ) \geqslant \frac{\epsilon}{2}
 \implies \frac{1}{n} \sum_{i=1}^{n}\! \left( \phi_\gamma(f_{\bar{g},\gamma}\left(z^\prime_i\right))\!-\!\phi_\gamma(f_{\bar{g},\gamma}\left(z_i\right)) \right) > \frac{\epsilon}{4}.
\end{align*}
This bounds the probability in step (1) as
\begin{align}
P^{2n}\left\{ \max_{\bar g \in \bar{\mathcal{G}}}\left( \frac{1}{n}\sum_{i=1}^n \left( \phi_{\gamma}\left(f_{\bar{g},\gamma}\left(Z^\prime_{i}\right)\right)- \phi_{\gamma}\left( f_{\bar{g},\gamma}\left(Z_i\right) \right) \right) \right)\! > \!\frac{\epsilon}{4} \right\}.\label{eq:prob-ready-sym}
\end{align}

3) For each $i$, $Z'_i$ and $Z_i$ admit the same distribution, and thus the difference in \eqref{eq:prob-ready-sym} is a symmetric random variable which allows one to do the second symmetrization by introducing independent random variables $\sigma_i$ taking values in $\{-1,1\}$ with equal probability. Then \eqref{eq:prob-ready-sym} is equal to
\begin{align}
\int_{\mathcal{Z}^{2n}}\! P_{\boldsymbol{\sigma}_n} \left( \max_{\bar g \in \mathcal{\bar G}} \frac{1}{n} \sum_{i=1}^{n}  \sigma_i  \left(\phi^{\prime}_{\bar g}\left(z^{\prime}_i\right)\!-\!\phi^{\prime}_{\bar g}\left(z_i\right) \right)\! >\! \frac{\epsilon}{4}\right)\!dP^{2n}(\mathbf{z}_{2n}) \label{eq:sec-symm}
\end{align}
where $\boldsymbol{\sigma}_n = \left( \sigma_i \right)_{1 \leqslant i \leqslant n}$.

4) Focusing on the integrand, apply the union bound and Hoeffding's inequality (Theorem~2 in \cite{Hoe63}), to upper bound the quantity \eqref{eq:sec-symm} by
$$
\exp\left(-\frac{n\epsilon^2}{32}\right)\int_{\mathcal{Z}^{2n}} \mathcal{N}\left(\frac{\epsilon\gamma}{8}, \mathcal{F}_{\mathcal{G},\gamma}, d_{1,\mathbf{z}_{2n}}\right) dP^{2n}\left(\mathbf{z}_{2n}\right).
$$
Finally, the claimed bound follows from the fact that the expected value of the covering number is less than $$\sup_{\mathbf{z}_{2n} \in \mathcal{Z}^{2n}}\mathcal{N}\left(\frac{\epsilon\gamma}{8}, \mathcal{F}_{\mathcal{G},\gamma}, d_{1,\mathbf{z}_{2n}}\right).$$
%give Inequality~\eqref{eq:uniform-convergence-L1}.
\end{proof}

\section{Bounds on Metric Entropy and Fat-shattering Dimension of Sets of BV Space} \label{sec:bv-space-metric-entropy}

%\subsection{Metric entropy}
The straightforwad way to estimate the $L_1$-norm metric entropy of sets of the $BV$ space is to appeal to Theorem~10.1.2 in \cite{Attouch14} which states the following. Let $\mathcal{T} \subset \mathbb{R}^d$. For any $f \in BV(\mathcal{T})$ and for any $\epsilon>0$, there exists a function $f_{\epsilon} \in C^{\infty}(\mathcal{T}) \cap W^{1,1}(\mathcal{T})=C^{\infty}(\mathcal{T}) \cap BV(\mathcal{T})$, such that 
$
\int_{\mathcal{T}}|f(t)-f_{\epsilon}(t)|dt < \epsilon,
$
and
$
||Df_{\epsilon}|(\mathcal{T})-|Df|(\mathcal{T})| < 4 \epsilon.
$
%where $|Df_{\epsilon}|(\mathcal{X})=\int_{\mathcal{X}}\lVert Df_{\epsilon}(x) \Vert_{2} dx$, $\lVert \cdot \rVert_2$ being the Euclidean norm.
Let $\mathcal{F} \subset BV(\mathcal{T})$ and suppose $W' \subset C^{\infty}(\mathcal{T}) \cap W^{1,1}(\mathcal{T})$ is the ball containing all $f_\epsilon$ satsifying the above conditions for each $f \in \mathcal{F}$. Suppose that $\bar{W}$ is an $\epsilon$-net of $W'$ with respect to the $L_1$-norm. Then, for any function $f_\epsilon \in W'$ there is a function $\bar{f} \in \bar{W}$, such that 
$$
\int_{\mathcal{T}}|f_{\epsilon}(t)-\bar{f}(t)|dt < \epsilon.
$$
By the triangle inequality, on the other hand,
\begin{align*}
\int_{\mathcal{T}}|f(t)-\bar{f}(t)|dt \leqslant \int_{\mathcal{T}}|f(t)-f_{\epsilon}(t)|dt + \int_{\mathcal{T}}|f_{\epsilon}(t)-\bar{f}(t)|dt < 2 \epsilon.
\end{align*}
This implies that $\bar{W}$ is a $2\epsilon$-net of $\mathcal{F}$. Then, according to Theorem~5.2 of \cite{Bir67}, the upper bound on the metric entropy of subsets of Sobolev spaces, it holds
$$
\ln \mathcal{N}\left(\epsilon, \mathcal{F}, d_{L_1}\right) \leqslant K \left(\frac{2}{ \epsilon}\right)^d,
$$
where $K$ is a constant possibly depending on $\mathcal{T}$ and $W'$. However, the explicit form of this dependency is not known which is clearly a donwside. 

Recently, \cite{Dutta18} derived an upper bound on the $L_1$-norm metric entropy of sets of the $BV$ space thanks to Poincar\'e type inequalities \cite{Attouch14}, with explicit constants. However, in view of Inequality~\eqref{eq:uniform-convergence-L1}, we need a weighted $L_1$-norm metric entropy estimate of the function class of interest (in fact, any weighted $L_p$-norm works thanks to Inequality~\eqref{eq:norm-ordering}). Using just the mentioned theorem, one can attempt at this as follows. Let $\mathcal{F} \subseteq BV(\mathcal{T}=[0,A]^d,[0,M])$ be of total variation $V$. If we assume $\mathcal{P}$ to be a family of probability distributions $P_{\mathcal{T}}$ on $\mathcal{T}$ with the Lebesgue density satisfying $\displaystyle{\left \lVert\frac{dP_{\mathcal{T}}}{dt} \right\rVert_{L_\infty} \leqslant K_{\mathcal{P}}}$ where $K_{\mathcal{P}} >0$, then from Theorem~3.1 in \cite{Dutta18} based on H\"older's inequality it follows that:
\begin{corollary} \label{cor:Dutta}
Fix $\epsilon \in (0, M]$. Then for any $P_{\mathcal{T}} \in \mathcal{P}$,
$$
\ln \mathcal{N}\left(\epsilon, \mathcal{F}, d_{L_1(P_{\mathcal{T}})}\right) \leqslant \frac{K M \left(\sqrt{d} A V K_{\mathcal{P}}\right)^d}{d K^2_{\mathcal{P}}}   \left(\frac{1}{\epsilon}\right)^d, \label{eq:}
$$
where $K$ is an absolute constant.
\end{corollary}
Now, we need the empirical version of the bound holding for $\displaystyle{\frac{1}{n} \sum_{i=1}^n \delta_{T_i}}$, a linear combination of Dirac measures supported on random variables $T_i$, $1 \leqslant i \leqslant n$, taking values in $\mathcal{T}$ and distributed independently according to $\mathcal{P}_{\mathcal{T}}$. One could do it based on the argument in Lemma~3 in \cite{BarKulPos97}. Following similarly to the proof of Theorem~\ref{theo:uniform-convergence-L1} and using Lemma~2 of \cite{BarKulPos97}, there holds
\begin{align*}
P^n_{\mathcal{T}} \left(\sup_{f,\bar{f} \in \mathcal{F}} \frac{1}{n} \sum_{i=1}^n\left|(f-\bar{f})(t_i)\right|\! -\! \int_{\mathcal{T}}\left|(f-\bar{f})(t)\right|dP_{\mathcal{T}} > \frac{\epsilon}{2}\right)\!\leqslant\!2 \mathcal{N}_1^2\left(\frac{\epsilon}{32},\mathcal{F},n\right)\exp\left(-\frac{n\epsilon^2}{64}\right),
\end{align*}
where $(f-g)(x)=f(x)-g(x)$.
Applying the combinatorial bound in \cite{MenVer03} (any other bound for general function classes such as Lemma~3.5 in \cite{AloBenCesHau97} could have been used, but this bound provides a better dependency on $\epsilon$),
\begin{align}
\mathcal{N}\left(\epsilon, \mathcal{F}, d_{2,\mathbf{t}_n}\right) \leqslant \left(\frac{7M}{\epsilon}\right)^{20d_{\mathcal{F}}\left(\frac{\epsilon}{96} \right)}, \label{eq:MenVer03}
\end{align}
to the right-hand side gives
\begin{align}
&P^n_{\mathcal{T}} \left(\sup_{f,\bar{f} \in \mathcal{F}} \frac{1}{n} \sum_{i=1}^n \left|(f-\bar{f})(t_i)\right| \!-\! \int_{\mathcal{T}}\left|(f-\bar{f})(t) \right|dP_{\mathcal{T}} > \frac{\epsilon}{2}\right) \nonumber \\ &\leqslant 2 \left(\frac{224M}{\epsilon}\right)^{40d_{\mathcal{F}}\left(\frac{\epsilon}{3072} \right)}\exp\left(-\frac{n\epsilon^2}{64}\right). \nonumber
\end{align}
Now, upper bound the right-hand side of the above inequality by $\delta \in (0,1)$. Let $\bar{\mathcal{F}}$ be an $\epsilon/2$-net of $\mathcal{F}$ with respect to the $L_1(P_{\mathcal{T}})$-norm. Then, for $$\displaystyle{n \geqslant K \left(\frac{d_{\mathcal{F}}\left(\epsilon/3072 \right)}{\epsilon^2}\ln \frac{1}{\epsilon} + \ln \frac{1}{\delta}\right)}$$ where $\delta$ is arbitrarily small, for almost all points $\mathbf{t}_n \in \mathcal{T}^n$, and for any $f \in \mathcal{F}$, there exists $\bar{f} \in \bar{\mathcal{F}}$ such that
\begin{align*}
\frac{1}{n} \sum_{i=1}^n \left|f(t_i)-\bar{f}(t_i)\right| - \int_{\mathcal{T}}\left|f(t)-\bar{f}(t) \right|dP_{\mathcal{T}} <  \frac{\epsilon}{2},
\end{align*}
implying 
\begin{align*}
\frac{1}{n} \sum_{i=1}^n \left|f(t_i)-\bar{f}(t_i)\right| < \epsilon.
\end{align*}
Then, an $\epsilon/2$-net of $\mathcal{F}$ with respect to the $L_1(P_{\mathcal{T}})$-norm, is an $\epsilon$-net of $\mathcal{F}$ with respect to the empirical $L_1$-norm. More precisely, 
\begin{align}
\ln \mathcal{N}\left(\epsilon, \mathcal{F}, d_{1,\mathbf{t}_n}\right) \leqslant \frac{K M \left(\sqrt{d} A V K_{\mathcal{P}}\right)^d}{d K^2_{\mathcal{P}}}   \left(\frac{2}{\epsilon}\right)^d, \label{eq:Dutta-empirical}
\end{align}
is the metric entropy of $\mathcal{F}$ with respect to the metric $d_{1,\mathbf{t}_n}$.
This bound however is inefficient in the sense that it holds (almost surely) for large values of $n$ and only for the distributions with Lebesgue densities. 

%\subsection{Fat-shattering dimension}
In fact, bounding the fat-shattering dimension of $\mathcal{F}$, then combining it with any metric entropy bound for general function classes (for instance the bound of \eqref{eq:MenVer03}) in the empirical metric, $d_{p,\mathbf{t}_n}$, would result in a dedicated (to the $BV$ space) metric entropy bound. In the following theorem, we bound the fat-shattering dimension of sets of the $BV$ space. 
\begin{theorem}\label{theo:metric-entropy-bounded-variation}
Fix $\epsilon \in (0, M]$. Then, there exists a positive constant $K$ such that
\begin{align}
d_{\mathcal{F}}(\epsilon) \leqslant \left(1+\frac{A \sqrt{VKd}}{\epsilon}\right)^d. \label{eq:fat-shattering-bounded-variation}
\end{align}
\end{theorem}
\begin{proof}
According to the fundamental result on line integrals (see, for instance \cite{Wil96}), and H\"older's inequality, for any $t_1,t_2 \in \mathcal{T}$, there holds  
\begin{align}
f(t_2)-f(t_1) &= \int_{0}^1 \langle Df(t_1+\beta(t_2-t_1)), t_2-t_1 \rangle d\beta \nonumber \\
&\leqslant \left(\int_{0}^1 \lVert Df(t_1+\beta(t_2-t_1)) \rVert^2_2 d\beta\right)^{\frac{1}{2}}  \left(\int_{0}^1 \lVert t_2 -t_1 \rVert^2_2 d\beta\right)^{\frac{1}{2}} \nonumber \\
&\leqslant \lVert t_2 -t_1 \rVert_2 \left(\int_{0}^1 \lVert Df(t_1+\beta(t_2-t_1)) \rVert^2_2 d\beta\right)^{\frac{1}{2}}. \label{eq:line-int}
%&\leqslant \lVert t_2 -t_1 \rVert_2 \lVert Df(t_0) \rVert_2,
\end{align}
where $\lVert \cdot \rVert_{2}$ is the standard Euclidean norm. Then, there exists $\phi \in \left(C^1_{c}(\mathcal{T}),\mathbb{R}^d \right)$ such that,
\begin{align*}
&\int_{0}^1 \lVert Df(t_1+\beta(t_2-t_1)) \rVert^2_2 d\beta \leqslant \int_{\mathcal{T}}\lVert Df(t) \rVert^2_2 dt \\
&=\int_{\mathcal{T}}\langle Df(t), Df(t) \rangle dt
\\
&\leqslant \int_{\mathcal{T}}\langle Df(t), \phi(t) \rangle dt \leqslant V K,
\end{align*}
for some constant $K$.
%The second and third inequalities are based on the fact that $\int_{\mathcal{T}}\langle Df(t), \phi(t) \rangle dt  \leqslant \left|Df \right|\left(\mathcal{T}\right) \lVert \phi \rVert_\infty $ for any $\phi \in \left(C^1_{c}(\mathcal{T}),\mathbb{R}^d \right)$. In fact, the second inequality becomes an equality if $f \in C_c^{1}(\mathcal{T})$.
Combining with \eqref{eq:line-int} gives
\begin{align}
f(t_2)-f(t_1) \leqslant \lVert t_2 -t_1 \rVert_2 \sqrt{V K} \label{eq:variation-sep}.
\end{align}
Now, suppose that $S=\{t_1, \dots, t_n\} \subset \mathcal{T}$ is a set of maximal cardinality $\epsilon$-shattered by $\mathcal{F}$. Re-arrange the indices so that $s(t_i) \leqslant s(t_{i+1})$. Since $S$ is $\epsilon$-shattered by $\mathcal{F}$, there exists a function $f$ in $\mathcal{F}$ satisfying $f(t_1) \leqslant s(t_1)-\epsilon$ and $f(t_i) \geqslant s(t_i)+\epsilon$ for all $i \neq 1$. Then, since $s(t_i)-s(t_1) \geqslant 0$,
$$
\forall t_i \in S \setminus \{t_1\}, \; f(t_i)-f(t_1) \geqslant s(t_i)-s(t_1)+2\epsilon \geqslant 2\epsilon.
$$
Thus, for any point $t_i$ with $i \leqslant n-1$ there exists a function $f$ in $\mathcal{F}$ for which $f(t_i) \leqslant s(t_i)-\epsilon$ and $f(t_j) \geqslant s(t_j)+\epsilon$ for all $\forall t_j \in S \setminus \{t_1,\dots,t_i\}$. Consequently,
$$
\forall t_j \in S \setminus \{t_1,\dots, t_i\}, \quad f(t_j)-f(t_i) \geqslant 2\epsilon.
$$
From these inequalities and from \eqref{eq:variation-sep} it follows that $S$ is $\displaystyle{\left(\frac{2\epsilon}{\sqrt{V K}}\right)}$-separated with respect to the Euclidean metric $d_2$. This implies that the fat-shattering dimension of $\mathcal{F}$ is at most the packing number of its domain:
$$ 
n \leqslant M\left(\mathcal{T},\left(\frac{2\epsilon}{\sqrt{VK}}\right),d_2\right).
$$
%. Let $\epsilon=\min_{x,x' \in S}d_2(x,x')$. 
Let $B$ be a unit ball in $\mathcal{T}$. By the volume comparison argument, it follows that
\begin{align*}
M(\mathcal{T},\epsilon,d_2) &\leqslant \frac{\left|\mathcal{T}+\frac{\epsilon}{2}B\right|}{\left|\frac{\epsilon}{2}B\right|} \leqslant \frac{\left|\sqrt{d} A B+\frac{\epsilon}{2}B\right|}{\left|\frac{\epsilon}{2}B\right|}
%&\leqslant \left(\frac{\sqrt{d} A+\epsilon/2}{\epsilon/2}\right)^d \frac{|B|}{|B|} \\
= \left(1+\frac{2\sqrt{d}A}{\epsilon}\right)^d,
\end{align*}
where $|A|$ denotes the volume of $A$.
Combining the two bounds gives the desired result.
\end{proof}
Now, substituting Inequality~\eqref{eq:fat-shattering-bounded-variation} in the combinatorial bound \eqref{eq:MenVer03}, yields:
\begin{align}
\ln \mathcal{N}\left(\epsilon, \mathcal{F}, d_{2,\mathbf{t}_n}\right) \leqslant 20 \left(\frac{198 A\sqrt{VKd}}{\epsilon}\right)^d \ln\left(\frac{7M}{\epsilon}\right). \label{eq:metric-entropy-bounded-variation}
\end{align}
In fact, one could use any metric entropy result for general function classes, such as Lemma~3.5 in \cite{AloBenCesHau97}
which depends on the sample size $n$:
\begin{align}
\ln \mathcal{N} \left(\epsilon, \mathcal{F}, d_{\infty, \mathbf{x}_n}\right) \!\leqslant\!d_{\mathcal{F}}\left(\frac{\epsilon}{4}\right) \log_2\!\left(\frac{2 M e n}{d_{\mathcal{F}}\left(\frac{\epsilon}{4}\right)\epsilon}\right)\!\ln\left(\frac{16 M^2 n}{\epsilon^2}\right)\!. \label{eq:AloBenCesHau97}
\end{align}
Notice that in both cases we have an additional logarithmic factor of $\epsilon^{-1}$, compared to the dedicated bound, Inequality~\eqref{eq:Dutta-empirical}, which at first sight might seem to be a drawback, particularly, for the latter bound displaying a $O(\ln^2\left(\epsilon^{-1}\right))$ as $\epsilon \rightarrow 0$. However, it will prove to be useful when elaborating the dependency on the number of classes which is addressed in the upcoming section.

\begin{remark}
Our bound on the fat-shattering dimension can be extended in a straightforward way to the $BV$ space on general metric spaces called doubling spaces, i.e., metric spaces where each ball can be covered by a finite number $k$ of balls of half the radius. This is possible thanks to the work of \cite{Mir03} extending the $BV$ functions to doubling spaces, and the packing number bound of \cite{Krau04} for sets of the mentioned spaces. In this case, the fat-shattering dimension will grow as a $O\left(\epsilon^{-\mbox{ddim}(\mathcal{X})}\right)$ as $\epsilon \rightarrow 0$, where $\mbox{ddim}(\mathcal{X})$ denotes the doubling dimension of $\mathcal{X}$, and is equal to $\log_2 k$. %The doubling dimension of $\mathbb{R}^d$ is roughly $d$.
\end{remark}

\section{Decomposition of Capacity Measures and Sample Complexity Estimate}\label{sec:sample-complexity}

In this section we estimate the sample size sufficient for the probability \eqref{eq:uniform-convergence-L1} to be at most $\delta \in (0,1)$ with the emphasis on making explicit the dependency of this estimate on the number of classes. The latter is possible thanks to decomposition result of a capacity measure of $\mathcal{F}_{\mathcal{G},\gamma}$ which upper bounds the mentioned quantity by that of $\mathcal{G}_0$. This can be done either at the level of the metric entropy or the fat-shattering dimension, since the former is related to the latter via combinatorial bounds (as has been seen in the preceding section).

\subsection{Metric Entropy}

For general function classes, the decomposition result for metric entropies in the $L_2(\mu)$-norm was provided in \cite{Duan12} and extended to all $L_p(\mu)$-norms in Lemma~1 in \cite{Gue17}. In the context of this work, it takes the following form:
\begin{align}
\ln \mathcal{N}\left(\epsilon, \mathcal{F}_{\mathcal{G},\gamma}, d_{p, \mathbf{z}_n}\right) \leqslant C \ln\mathcal{N} \left(\frac{\epsilon}{C^{\frac{1}{p}}}, \mathcal{G}_0, d_{p, \mathbf{x}_n} \right). \label{eq:decom-metric-entropy}
\end{align}
The important part in this bound to pay attention to is the scale of the metric entropy of the component class on the right-hand side which depends on the number of classes: the ''worst`` case is $p=1$ giving $\epsilon /C$ dependency and the ''optimal`` case corresponds to $p=\infty$ in which case the dependency on $C$ vanishes. Applying it to the metric entropy bound \eqref{eq:Dutta-empirical} with $p=1$, or to \eqref{eq:metric-entropy-bounded-variation} with $p=2$, would yield a result scaling with $C$ as a $O(C^{d+1})$.
 %Because $C$ appears in the scale of covering numbers in \eqref{eq:decom-metric-entropy}, it gets passed to the scale of the fat-shattering dimension which grows as a $\displaystyle{O\left(\frac{1}{\epsilon^d}\right)}$, and as the
%In view of Theorem~\ref{theo:metric-entropy-bounded-variation}, instantiating Inequality \eqref{eq:decom-metric-entropy} in $p=2$ and applying the mentioned theorem to the right-hand side gives
%\begin{align}\ln \mathcal{N}\left(\epsilon, \mathcal{F}_{\mathcal{G},\gamma}, d_{p, \mathbf{z}_n}\right) \!\leqslant \! 20 C^{\frac{d}{2}+1} \left(\frac{198 \sqrt{dV} A }{\epsilon}\right)^d \ln\left(\frac{7M C}{\epsilon}\right).\label{eq:entropy-FG-decom-cov}\end{align}
 %Interestingly, this would also be the case if we applied the decomposition \eqref{eq:decom-metric-entropy} with $p=1$ to the dedicated metric entropy bound \eqref{eq:Dutta-empirical}, instead of Theorem~~\ref{theo:metric-entropy-bounded-variation}, as the former exhibits a $\displaystyle{O\left(\frac{1}{\epsilon^d}\right)}$ dependency. 
 For problems involving a large number of classes and high dimensional input spaces, this is quite a prohibitive dependency. Now, for $2 <p < \infty$, we can use the metric entropy bounds established in Corollary~1 in \cite{MusLauGue19} which when applied to sets of $BV$ gives results scaling with $C$ as
\begin{align}
&\ln \mathcal{N} \left(\epsilon, 
\mathcal{F}_{\mathcal{G}, \gamma}, d_{p, \mathbf{z}_n}\right) \leqslant 2 C \log^d_2(2C)\left(\frac{60A\sqrt{VKd}}{\epsilon}\right)^d
\ln\left(\frac{30en\log_2\left(2C\right) M}{\epsilon}\right), \nonumber%\label{eq:cor1Mus19}
\end{align}
and slightly worse for the dimension-free one, but still an improvement over the cases $p \in \{1,2\}$. Now, contrast it with the extreme case $p=\infty$ for which we have
%. Notice how, in this case, $C$ disappears from the scale of the metric entropy of $\mathcal{G}_0$:
\begin{align}
\ln \mathcal{N}\left(\epsilon, \mathcal{F}_{\mathcal{G},\gamma}, d_{\infty, \mathbf{z}_n}\right)
\leqslant C \ln\mathcal{N} \left(\epsilon, \mathcal{G}_0, d_{\infty, \mathbf{x}_n} \right), \label{eq:decom-metric-entropy-ext}
\end{align}
and combining it with \eqref{eq:AloBenCesHau97} leads to 
\begin{align}
&\ln \mathcal{N} \left(\epsilon, \mathcal{F}_{\mathcal{G},\gamma}, d_{\infty, \mathbf{z}_n}\right) \nonumber \\
&\leqslant C d_{\mathcal{G}_0}\left(\frac{\epsilon}{4}\right)  \log_2\left(\frac{2 M e n}{d_{\mathcal{G}_0}\left(\frac{\epsilon}{4}\right)\epsilon}\right) \ln\left(\frac{16 M^2 n}{\epsilon^2}\right). \label{eq:AloBenCesHau97-decom}
\end{align}
All we need to do now is to substitute Inequality \eqref{eq:fat-shattering-bounded-variation} on $d_{\mathcal{G}_0}\left(\epsilon\right)$ in the above result which will not affect the dependency on $C$. We now have a $O(C)$ dependency compared to a $O(C^{{d}+1})$ and a $O(C\ln^{d+2}(C))$. However, the price to pay for this improvement is the dependency of Inequality \eqref{eq:AloBenCesHau97-decom} on the sample size $n$ as a $O(\ln^2(n))$. In the following section, we provide a new efficient decomposition result for the fat-shattering dimension. This leads to a new metric entropy bound for $\mathcal{F}_{\mathcal{G},\gamma}$ in the empirical $L_2$-norm.

\subsection{Fat-shattering Dimension}

Inspired from the work of \cite{Vid13} concerning the decomposition of the classical VC-dimension, Theorem~6.2 in \cite{Duan12} provides that for the fat-shattering dimension which  in the context of this work gives:
\begin{align}
d_{\mathcal{F}_{\mathcal{G}}}\left(\epsilon\right) \leqslant 462 C d_{\mathcal{G}_{0}}\left(\frac{\epsilon}{96\sqrt{C}}\right) \ln \left(\frac{24 M_{\mathcal{G}} \sqrt{C}}{\epsilon}\right). \label{eq:duan-fat-decom}
\end{align}
Notice again that the scale of the fat-shattering dimension of the component class $\mathcal{G}_0$ depends on $C$, and thus in view of a $\displaystyle{O\left(\epsilon^{-d}\right)}$ dependency established in \eqref{eq:fat-shattering-bounded-variation} will yield a $O(C^{\frac{d}{2}+1})$ dependency on $C$. We provide an amelioration over it thanks to the following new lemma relating the fat-shattering dimension to the empirical $L_\infty$-norm metric entropy which is interesting on its own right.
\begin{lemma} \label{lemma:from-fat-to-entropy}
Let $\mathcal{F}$ be a uniformly-bounded class of real-valued functions on some metric space $\mathcal{T}$. Denote $d=d_{\mathcal{F}}(\epsilon)$. Then,
$$
d \leqslant \log_{2}\mathcal{N}_{\infty}\left(\epsilon,\mathcal{F},d\right).
$$
\end{lemma}
\begin{proof}
Fix $\epsilon>0$. Let $\mathcal{T}_d \subset \mathcal{T}$ be the set of maximal cardinality $d$ $\epsilon$-shattered by $\mathcal{F}$. Then, we can distinguish a subset $\mathcal{F}^{\prime}$ in $\mathcal{F}$ such that for any pair of different functions $f$ and $f'$ in $\mathcal{F}^{\prime}$, there is a point $t$ in $\mathcal{T}_d$ such that $\left|f(t)-f^{\prime}(t)\right| \geqslant 2\epsilon$ and consequently $ \max_{t \in \mathcal{T}_d} \left|f(t)-f^{\prime}(t)\right| \geqslant 2\epsilon $. The cardinality of this set is at least $2^d$. This implies that,
$$ 2^{d} \leqslant \mathcal{M} \left(2\epsilon, \mathcal{F}, d_{\infty,\mathbf{t}_d} \right) \leqslant \mathcal{M}_{\infty} \left(2\epsilon, \mathcal{F}, d \right),$$ and the claimed bound follows from $ \mathcal{M}_{\infty} \left(2\epsilon, \mathcal{F}, d \right) \leqslant \mathcal{N}_{\infty}\left(\epsilon, \mathcal{F}, d \right).$
\end{proof}
\begin{remark}
The immediate consequence of this result is that it may improve upon the known estimates on the fat-shattering dimension such as that of the following function classes. %provide more efficient estimates on the fat-shattering dimension of the following class of functions. 
If according to \cite{Asor14} the fat-shattering dimension of the class of the sine functions on the bounded interval grows as a $O(\epsilon^{-1})$ as $\epsilon \rightarrow 0$, then from Lemma~\ref{lemma:from-fat-to-entropy} and Inequality~177 in \cite{KolTih61} (which concerns a class of entire functions of which sine functions are member) one obtains a $O(\ln\epsilon^{-1})$ dependency. On the other hand, for sets in Gaussian reproducing kernel Hilbert spaces, Lemma~\ref{lemma:from-fat-to-entropy} and Lemma~4.5 in \cite{VaaZan09} leads to a bound comparable to Theorem~12 of \cite{Belkin18}, but with an explicit dependency on the ''width`` of the kernel, the parameter characterizing the ''complexity`` of the function class.
\end{remark}

The fact that we are now dealing with the metric entropy in the uniform metric is quite convenient because of the form of dependency on $C$ on the right-hand side of \eqref{eq:decom-metric-entropy-ext}. Our decomposition of the fat-shattering dimension of $\mathcal{F}_{\mathcal{G},\gamma}$ is then as follows:
\begin{theorem} \label{theo:decom-fat}
Fix $\epsilon \in (0,M]$ and let $d=d_{\mathcal{F}_{\mathcal{G},\gamma}}(\epsilon)$. Then,
\begin{align}
d &\leqslant 32 Cd_{\mathcal{G}_{0}}\left(\frac{\epsilon}{4}\right) \log^2\left(\frac{256 C M^2}{\epsilon^2} d_{\mathcal{G}_{0}}\left(\frac{\epsilon}{4}\right) \right).
 \label{eq:fat-covering-decom}
\end{align}
\end{theorem}
\begin{proof}
Apply Lemma~\ref{lemma:from-fat-to-entropy} to $\mathcal{F}_{\mathcal{G},\gamma}$ and the decomposition formula \eqref{eq:decom-metric-entropy} to get
\begin{align}
d \leqslant C \log_2 \mathcal{N}_\infty \left(\epsilon, \mathcal{G}_0, d \right). \label{eq:fat-covering-decomp-prefinal}
\end{align}
Combining with \eqref{eq:AloBenCesHau97} and assuming that $d_{\mathcal{G}_{0}}\left(\frac{\epsilon}{4}\right) \geqslant 1$, it holds
%\begin{align*}
%d &\leqslant C d_{\mathcal{G}_{0}}\left(\frac{\epsilon}{4}\right) \log_2\left(\frac{2 M e d}{d_{\mathcal{G}_0}\left(\frac{\epsilon}{4}\right)\epsilon}\right) \log_2\left(\frac{16 M^2 d}{\epsilon^2}\right).
%\end{align*}
\begin{align*}
d &\leqslant C d_{\mathcal{G}_{0}}\left(\frac{\epsilon}{4}\right) \log^2_2\left(\frac{16 M^2 d}{\epsilon^2}\right).
\end{align*}
Next, we use the following result which appears as a partial result in the proof of Theorem~17 in \cite{Bar98}: for any $a,b \geqslant 1$ and for any $x \geqslant 1$,
$$a \log^2_2\left(b x\right) \leqslant \frac{x}{2}+16 a \log^2_2\left(16 a b \right).$$
As a result, we have
\begin{align*}
d &\leqslant \frac{d}{2} + 16Cd_{\mathcal{G}_{0}}\left(\frac{\epsilon}{4}\right) \log^2\left(\frac{ 256 C M^2}{\epsilon^2} d_{\mathcal{G}_{0}}\left(\frac{\epsilon}{4}\right) \right)
\end{align*}
and the desired bound follows.
\end{proof}
\begin{remark}
Yet another possibility for the decomposition of the fat-shattering dimension is via the (empirical) Rademacher complexity. Using the decomposition results for the Rademacher complexity of general function classes such as the one in \cite{Mau16} or \cite{KuzMohSye14} with a linear dependency on $C$, and taking into account that the Rademacher complexity of most function classes on the unit ball of their domains is upper bounded by $K_{\mathcal{F}}/\sqrt{n}$ (see, for instance, \cite{BarMen02}), where $K_{\mathcal{F}}$ is the quantity characterizing the class $\mathcal{F}$, it follows that
$$
\hat{R}_n(\mathcal{F}) \leqslant \frac{CK_{\mathcal{F}}}{\sqrt{n}}.
$$
Then, according to Lemma~\ref{lemma:mendelson-fat-shattering} in Appendix \ref{app:fat-rad},
$$d_{\mathcal{F}}(\epsilon) \leqslant \frac{C^2K^2_{\mathcal{F}}}{\epsilon^2}.$$
This exhibits a worse dependency on $C$ than the one established in Theorem~\ref{theo:decom-fat}.
\end{remark}
We now apply the combinatorial bound \eqref{eq:MenVer03} to the class $\mathcal{F}_{\mathcal{G},\gamma}$, then use Theorem~\ref{theo:decom-fat} to obtain:
\begin{corollary} \label{cor:new-bound}
For any $\epsilon \in (0,\gamma]$ and any $n > 0$
\begin{align}
&\ln \mathcal{N}\left(\epsilon, \mathcal{F}_{\mathcal{G},\gamma}, d_{2,\mathbf{z}_n}\right) \leqslant 640 C d_{\mathcal{G}_{0}}\left(\frac{\epsilon}{384}\right) \ln^2\left(\frac{256 C M^2}{\epsilon^2} d_{\mathcal{G}_{0}}\left(\frac{\epsilon}{384}\right) \right)\ln\left(\frac{7\gamma}{\epsilon}\right). \label{eq:entropy-FG-decom-fat}
\end{align}
\end{corollary}
This result provides an improvement over Corollary~1 in \cite{MusLauGue19} that we mentioned in the preceding subsection in terms of the dependency on $C$ (since, now, $C$ does not appear inside the scale of the component fat-shattering dimension), as well as over the bound \eqref{eq:AloBenCesHau97-decom} in terms of the dependency on the sample size. 
%It is also interesting to compare this bound with \eqref{eq:AloBenCesHau97-decom}: there is no dependency on the sample size but instead, 
But the price to pay for such an improvement is  having $C$ and the component fat-shattering dimension appearing inside a logarithmic factor as well as an additional $\displaystyle{\ln\left(\frac{7\gamma}{\epsilon}\right)}$ factor.

\subsection{Sample Complexity}

The following result shows the difference between the sample complexity estimates obtained based on these two metric entropy bounds, \eqref{eq:AloBenCesHau97-decom} and \eqref{eq:entropy-FG-decom-fat}. Although the two metric entropy bounds used exhibit different dependencies on the number of classes, both sample complexity results scale as a $O(C \ln^2 (C))$. Also, one can see that the sample-size free bound (which corresponds to the second result) does not provide any better sample complexity estimate. 
\begin{theorem}\label{theo:sample-complexity-iid}
Fix $\epsilon, \delta \in (0,1)$ and fix $\gamma \in (0, 1]$. Let $\mathcal{G}=\mathcal{G}_0^C$ where $\mathcal{G}_0 \subseteq BV([0,A]^d, [0,M])$. Let $\displaystyle{F=\left(\frac{A\sqrt{VKd}}{\epsilon \gamma}\right)^d}$. Then, for a sample size at least
$$\frac{1}{\epsilon^2}\left(K_1 C F \ln^2\left(\frac{CM^2}{\epsilon^2\gamma^2}F \right)+ \ln \frac{2}{\delta}\right)$$ obtained via \eqref{eq:AloBenCesHau97-decom}
and 
$$
\frac{1}{\epsilon^2}\left(K_2C F \ln^2\left(\frac{C M^2}{\epsilon^2 \gamma^2} F \right)\ln\left(\frac{1}{\epsilon}\right) + \ln{\frac{2}{\delta}}\right)
$$ obtained via \eqref{eq:entropy-FG-decom-fat},
%$$
%	n \geqslant \frac{32}{\epsilon^2} \times \begin{cases} 
%		\displaystyle{K_1 C d\left(\frac{\epsilon \gamma}{32}\right)\ln^2\left(\frac{CM^2}{\epsilon^2\gamma^2}d\left(\frac{\epsilon \gamma}{32}\right)\right)+ \ln \frac{2}{\delta}}, &\mbox{via \eqref{eq:AloBenCesHau97-decom}}, \\
%	\displaystyle{K_2C d\left(\frac{\epsilon \gamma}{3072}\right) \ln^2\left(\frac{C M^2}{\epsilon^2 \gamma^2} d\left(\frac{\epsilon  \gamma}{3072}\right) \right)\ln\left(\frac{1}{\epsilon}\right) + \ln{\frac{2}{\delta}}}, &\mbox{via \eqref{eq:entropy-FG-decom-fat}}; 
%	\end{cases}
%$$
where $0<K_1<K_2$ are constans, the probability \eqref{eq:uniform-convergence-L1} is at most $\delta$. 
\end{theorem}
\begin{proof}
For the second bound, apply Inequality \eqref{eq:entropy-FG-decom-fat} to the right-hand side of Inequality~\ref{eq:uniform-convergence-L1} to bound it by
\begin{align*}
%&P^{n}\left(\sup_{g \in \mathcal{G}} \left( L_{\gamma}\left(g\right)- L_{\gamma, n}\left(g\right) \right)  > \epsilon \right)  \\ & \leqslant
2&\exp\left(640 C d\left(\frac{\epsilon \gamma}{3072}\right)  \ln^2\left(\frac{256 C M^2}{\epsilon^2 \gamma^2} d\left(\frac{\epsilon \gamma}{3072}\right)  \right)\ln\left(\frac{56}{\epsilon}\right)-\frac{n\epsilon^2}{32}\right).
\end{align*}
Upper bound the right-hand side by $\delta$ and solve for $n$. The proof of the first bound proceeds similarly in which case to solve for $n$ we make use of the inequality
$$\ln n \leqslant \sqrt{Kn}+\ln\left(\frac{4}{Ke^2}\right),$$
where we set $K=\epsilon^2/\left(C d\left(\epsilon \gamma/32\right)\right)$.
\end{proof}
%Thus we conclude that in this context the bound \eqref{eq:AloBenCesHau97-decom} has an advantage over the dedicated bound \eqref{eq:Dutta-empirical}  in terms of the dependency on $C$ as well as \eqref{eq:entropy-FG-decom-fat} in terms of the scale of the fat-shattering, the dependency on $\epsilon$ and the constants. However, the entropy bound \eqref{eq:entropy-FG-decom-fat} for the class $\mathcal{F}_{\mathcal{G},\gamma}$ might prove to be useful in other settings based on the way it depends on $C$ and non-dependency on the sample size, an improvement compared to Corollary~1 in \cite{MusLauGue19} where $C$ appears inside the scale of the component fat-shattering dimension.

\section{Conclusions} \label{sec:conclusions}
The present paper dealt with the derivation of the minimal sample size estimate sufficient for the empirical and generalization performances to be very close with high probability for multi-category classifiers when 1) these performances are assessed based on the truncated hinge loss function, and 2) the functions they implement are of bounded variation defined on $\mathbb{R}^d$. We were particularily interested in elaborating the dependency of the sample size estimate on the number $C$ of classes. To this end, first we generalized the uniform deviation result of \cite{BarLon95} to the multi-category setting. Second, we upper bounded the fat-shattering dimension of classes of $BV$ functions which gave a result scaling as a $O(\epsilon^{-d})$ as $\epsilon \rightarrow 0$. This can be substituted in a combinatorial bound for general function classes giving a dedicated to the $BV$ space bound. Concerning the dependency on $C$, we appealed to a particular bound, the decomposition of capacity measure, and improved upon the known decomposition of the fat-shattering dimension \cite{Duan12}: our result scales as a $O(C \ln^2 C)$ compared to a $O(C^{\frac{d}{2}+1})$ for the $BV$ sets. Using our result then gives a sample complexity estimate with the same dependency on $C$ improving upon those obtained based on the decomposition of the empirical $L_p$ norms with $1 \leqslant p < \infty$, and comparable to the one in $p=\infty$. 

So far the decomposition of the fat-shattering dimension was done via other capacity measures: the metric entropy and the Rademacher complexity. Although our bound obtained via the metric entropy demonstrates a tighter dependency on $C$, it involves a $\ln^2(\epsilon^{-1})$ factor (inherited from the combinatorial bound of \cite{AloBenCesHau97}), a deterioration compared to the decomposition via a Rademacher complexity. From the argument made in \cite{Men02} it seems that the direct decomposition (by not appealing to any intermediate capacity measure) of the fat-shattering dimension, might provide a dependency on $C$ closer to the case involving the Rademacher complexity. In this sense, we aim to take benefit from the bounded variation assumption---which is not too restrictive---to improve the dependency on $C$ of the decomposition of the Rademacher complexity, and thus that of the fat-shattering dimension.

\appendix

\section{Non independent case: stationary $\beta$-mixing data} \label{app:beta-mixing}
Assuming that the data is distributed in an independent fashion is rather restrictive, since many real-world problems (the classical example are problems dealing with time series data) fail to satisfy it. Here we consider a setting which slightly weakens the independence assumption yet renders itself amenable to the tools from the VC framework: this setting is that of mixing processes \cite{Bra86}. Below, we first give the definition of stationary $\beta$-mixing process and extend Theorem~\ref{theo:uniform-convergence-L1} to such sequences, then, we derive (effective) sample complexity result.

\subsection*{Uniform convergence for stationary $\beta$-mixing data}
%We consider multi-category pattern classification problems where the goal is to assign an object represented by $x \in \mathcal{X}$ (description space) to one of the categories $y$ in $\mathcal{Y}=\{ 1, \dots, C \}$ with $2 < C < \infty$. Let $\mathcal{Z}=\mathcal{X} \times \mathcal{Y}$, and $\mathcal{A}=\mathcal{A}_{\mathcal{X}} \times \mathcal{A}_{\mathcal{Y}}$, a product of sigma algebras $\mathcal{A}_{\mathcal{X}}$ and $\mathcal{A}_{\mathcal{Y}}$ defined on $\mathcal{X}$ and $\mathcal{Y}$, respectively.  %Denote by $X, Y$ random variables taking values in $\mathcal{X}$ and $\mathcal{Y}$, respectively. 
Keeping the same notation as in the main text, we assume that the sequence $\mathbf{Z}=(Z_i=(X_i,Y_i))_{i \geqslant 0}$ of random variables defined on the product space $(\mathcal{Z}^{\infty},\mathcal{A}^{\infty}, \mathbb{P})$ is strictly stationary: for any $l \geqslant 0$, $(Z_{i_1}, \dots, Z_{i_k})$ and $(Z_{i_1+l}, \dots, Z_{i_k+l})$ admit the same probability structure which, in a particular case, implies that $Z_i$ and $Z_j$ for any $i,j$ have the same distribution. Let $\mathbf{Z}_{k,l}=(Z_i)_{k \leqslant i \leqslant l}$. Let $\sigma(\mathbf{Z}_{k,l})$ denote the sigma-algebra generated by the sequence $\mathbf{Z}_{k,l}$. Fix $k > 0$. $\mathbf{Z}$ is said to be $\beta$-mixing (or absolutely/completely regular) if the quantity 
$$\beta(k)=\sup_{l \geqslant 1}\mathbb{E}_{A \in \sigma(\mathbf{Z}_{1,l})} \left[\sup_{A' \in \sigma(\mathbf{Z}_{l+k,\infty}) } |\mathbb{P}\left( A' | A \right) - \mathbb{P}\left(A'\right)|\right].$$
goes to zero as $k \rightarrow \infty$. There exist weaker as well as stronger mixing settings than just the defined one. %, which are $\alpha$ and $\phi$-mixing, respectively. 
%If the former constitutes weaker mixing conditions, the latter is stronger than $\beta$-mixing, 
 If the results stated below holds true in the stronger than $\beta$-mixing setting, to the best of our knowledge, it is not yet clear if the uniform law of large numbers holds for the weaker mixing processes (the law of large numbers for such processes is proved in \cite{Vid13}). 
 
The result below, when $\beta$-mixing coefficient vanishes, reduces to the result in the i.i.d. case, albeit with a slower convergence rate.
\begin{theorem}
Suppose that the sequence $\mathbf{Z}_n=(Z_i)_{1 \leqslant i \leqslant n}$ is drawn from a stationary $\beta$-mixing distribution. Divide it into $2b_n$ blocks each of size $a_n\geqslant 1$  $2b_n a_n =n$. Fix $\epsilon \in (0,1)$ and $\gamma \in (0, 1]$. Then for any $b_n>\frac{2}{\epsilon^2 a_n}$, there holds
\begin{align}
\mathbb{P}\left(\sup_{g \in \mathcal{G}} \left( L_{\gamma}\left(g\right)- L_{\gamma, n}\left(g\right) \right)  > \epsilon \right) \leqslant 4 \exp\left( -\frac{b_n  \epsilon^2}{32}\right) \mathcal{N}_1\left(\frac{\epsilon \gamma}{16},\mathcal{F}_{\mathcal{G},\gamma}, 2n\right) +  2b_n \beta (a_n). \label{eq:lala}
\end{align}
%\begin{align}
%\mathbb{P} \left(\sup_{f \in \mathcal{F}} L(f) - L_n(f) > \epsilon \right) &\leqslant 4 \exp\left( -\frac{\mu_n  \epsilon^2}{32}\right) \mathcal{N}_1\left(\frac{\epsilon L}{16},\mathcal{F}, 2n\right) +  4\mu_n \beta (a_n). \label{eq:lala}
%\end{align}
\end{theorem}
 The proof follows the approach in \cite{Meir2000} which focuses on time series prediction, as well as that of Theorem~\ref{theo:uniform-convergence-L1}. The sensitive parts of the proof of the latter theorem to the independence assumption are concentration inequalities: they can be readily applied by passing from the original sequence to a sequence of independent blocks. In the proof, to ease the reading, $\mathbb{E}_{P} X$ is used instead of $\mathbb{E}_{X \sim P} X$.
 %expectation taken with respect to the probability measure $P$.
\begin{proof}
For $j \in \{1, \dots, b_n \}$, let $S_j=\{ i : 2(j-1)a_n+1 \leqslant i \leqslant (2j-1)a_n\}$ and $S'_j=\{ i : (2j-1)a_n+1 \leqslant i \leqslant 2ja_n\}$. Denote $\mathbf{Z}^{(j)}=(Z_i)_{i \in S_j}$ and $\mathbf{Z}_{a_nb_n}=\left(\mathbf{Z}^{(j)}\right)_{1 \leqslant j \leqslant b_n}$. Let $\mathbf{\tilde{Z}}_{a_n b_n}=\left(\mathbf{\tilde{Z}}^{(j)}\right)_{1 \leqslant j \leqslant b_n}$ be a sequence of blocks $\mathbf{\tilde{Z}}^{(j)}=(\tilde{Z}_i)_{i \in S_j}$ independent from $\mathbf{Z}_{a_n b_n}$, each of which is distributed independently according to the marginal distribution of the original blocks $\mathbf{Z}^{(j)}$.
%but independent of $\mathbf{Z}^{(j)}$ for any $j \in \{1, \dots, \mu_n \}$.
Thanks to the stationarity property, all blocks $\mathbf{Z}^{(j)}$ (as well as the blocks $\mathbf{Z}^{'(j)}=(Z_i)_{i \in S'_j}$) have the same marginal distribution which we denote by $\mathbb{P}_{a_n}$. Thus that of $\mathbf{\tilde{Z}}_{a_n b_n}$  is $\mathbb{P}^{b_n}_{a_n}$ ($b_n$ times $\mathbb{P}_{a_n}$). 

Let $\mathbb{Q}$ be the distribution of $\mathbf{Z}_{a_n b_n}$. %and let $\tilde{\mathbb{Q}}=\mathbb{P}^{b_n}_{a_n}$.
Thanks to Lemma~4.1 in \cite{Yu94} the sequence $\mathbf{Z}_{a_n b_n}$ can be related to $\mathbf{\tilde{Z}}_{a_n b_n}$ via 
\begin{align}
\left|\mathbb{E}_{\mathbb{Q}} f(\mathbf{Z}_{a_n b_n}) - \mathbb{E}_{\mathbb{P}^{b_n}_{a_n}} f(\mathbf{\tilde{Z}}_{a_n b_n})\right| \leqslant b_n \beta (a_n) \left\lVert f \right\rVert_{\infty}, \label{eq:from-dep-to-indep}
\end{align}
where $f$ is a bounded measurable function on $\mathcal{Z}^{b_n a_n}$. Let $\mathcal{F}$ denote a set of such functions. Now, let $\mathbb{P}_0$ be the one-dimensional marginal of $\mathbb{P}$ and let $f_{S}(\mathbf{Z}^{(j)})=\sum_{i \in S_j} f(Z_i)$ and $\tilde{f}_{S}(\mathbf{\tilde{Z}}^{(j)})=\sum_{i \in S_j} f(\tilde{Z}_i)$. It follows that
\begin{align*}
&\mathbb{P} \left(\sup_{f \in \mathcal{F}} \mathbb{E}_{\mathbb{P}_0} f(Z) - \frac{1}{n} \sum_{i=1}^n f(Z_i) > \epsilon \right) \\
&= \!
\mathbb{P}\! \left(\!\sup_{f \in \mathcal{F}} \frac{1}{2a_n} \left(\mathbb{E}_{\mathbb{P}_{a_n}}\sum_{i \in S_1} f(Z_i)\!+\!\mathbb{E}_{\mathbb{P}_{a_n}}\sum_{i \in S'_1} f(Z_i) \right)\!-\! \left(\!\frac{1}{n} \sum_{j=1}^{b_n} \!\left(\!\sum_{i \in S_j} f(Z_i)+ \sum_{i \in S'_j} f(Z_i) \right)\right)\! > \!\epsilon \right) \\
%&\leqslant 
%\mathbb{P} \left(\sup_{f \in \mathcal{F}} \left(\frac{1}{2a_n} \mathbb{E}_{\mathbb{P}_{a_n}}\sum_{i \in S_1} f(Z_i) - \frac{1}{2b_n a_n} \sum_{j=1}^{b_n} \sum_{i \in S_j} f(Z_i)\right)+ \sup_{f \in \mathcal{F}}\left(\frac{1}{2a_n} \mathbb{E}_{\mathbb{P}_{a_n}}\sum_{i \in S'_1} f(Z_i) -\frac{1}{2b_n a_n} \sum_{j=1}^{b_n} \sum_{i \in S'_j} f(Z_i) \right) > \epsilon \right) \\
&\leqslant 
\mathbb{P} \left(\sup_{f \in \mathcal{F}}\frac{1}{2a_n} \mathbb{E}_{\mathbb{P}_{a_n}}\sum_{i \in S_1} f(Z_i) - \frac{1}{2b_n a_n} \sum_{j=1}^{b_n} \sum_{i \in S_j} f(Z_i) > \frac{\epsilon}{2} \right) \\
&+ \mathbb{P} \left(\sup_{f \in \mathcal{F}}\frac{1}{2a_n} \mathbb{E}_{\mathbb{P}_{a_n}}\sum_{i \in S'_1} f(Z_i) -\frac{1}{2b_n a_n} \sum_{j=1}^{b_n} \sum_{i \in S'_j} f(Z_i)> \frac{\epsilon}{2} \right) \\
& \leqslant 
2 \mathbb{P} \left(\sup_{f \in \mathcal{F}}\mathbb{E}_{\mathbb{P}_{a_n}}\sum_{i \in S_1} f(Z_i) - \frac{1}{b_n} \sum_{j=1}^{b_n} \sum_{i \in S_j} f(Z_i) > a_n \epsilon \right) \\
&=2 \mathbb{P} \left(\sup_{f \in \mathcal{F}}\mathbb{E}_{\mathbb{P}_{a_n}}f_{S}(\mathbf{Z}^{(1)}) - \frac{1}{b_n} \sum_{j=1}^{b_n}f_{S}(\mathbf{Z}^{(j)}) > a_n \epsilon \right),
\end{align*}
where the first equality follows from the linearity of the expectation and the stationarity property, the first inequality from the sub-additivity of the supremum and the union bound, and the second one from the stationarity property. From \eqref{eq:from-dep-to-indep} it follows
\begin{align*}
&\mathbb{P} \left(\sup_{f \in \mathcal{F}}\mathbb{E}_{\mathbb{P}_{a_n}}f_{S}(\mathbf{Z}^{(1)}) - \frac{1}{b_n} \sum_{j=1}^{b_n}f_{S}(\mathbf{Z}^{(j)}) > a_n \epsilon \right) \\ &\leqslant \mathbb{P}^{b_n}_{a_n} \left(\sup_{f \in \mathcal{F}}\mathbb{E}_{\mathbb{P}_{a_n}}\tilde{f}_{S}(\mathbf{\tilde{Z}}^{(1)}) - \frac{1}{b_n} \sum_{j=1}^{b_n}\tilde{f}_{S}(\mathbf{\tilde{Z}}^{(j)}) > a_n \epsilon \right) + b_n \beta (a_n),
\end{align*}
and thus
\begin{align*}
\mathbb{P} \left( \sup_{f \in \mathcal{F}} \mathbb{E}_{\mathbb{P}_0} f(Z) - \frac{1}{n} \sum_{i=1}^n f(Z_i) > \epsilon \right) &\leqslant 2\mathbb{P}^{b_n}_{a_n} \left(\sup_{f \in \mathcal{F}}\mathbb{E}_{\mathbb{P}_{a_n}}\tilde{f}_{S}(\mathbf{\tilde{Z}}^{(1)}) - \frac{1}{b_n} \sum_{j=1}^{b_n}\tilde{f}_{S}(\mathbf{\tilde{Z}}^{(j)}) > a_n \epsilon \right) \\
&+  2b_n \beta (a_n).
\end{align*}
For any $f_{g,\gamma} \in \mathcal{F}_{\mathcal{G},\gamma}$, let $\tilde{l}_{f_{g,\gamma}}: \mathcal{Z}^{a_n} \rightarrow \mathbb{R}$ be defined as $\tilde{l}_{f_{g,\gamma}}(\mathbf{Z}_{a_n})=\sum_{i=1}^{a_n}\phi_\gamma(f_{g,\gamma}(Z_i))$. Let $\tilde{L}=\left\{ \tilde{l}_{f_{g,\gamma}} : f_{g,\gamma} \in \mathcal{F}_{\mathcal{G},\gamma}\right\}$. Applying the derivations above to $\mathbb{P}\left(\sup_{g \in \mathcal{G}} \left( L_{\gamma}\left(g\right)- L_{\gamma, n}\left(g\right) \right)  > \epsilon \right)$, we get
\begin{align*}
\mathbb{P}\left(\sup_{g \in \mathcal{G}} \left( L_{\gamma}\left(g\right)- L_{\gamma, n}\left(g\right) \right)  > \epsilon \right) &\leqslant 2\mathbb{P}^{b_n}_{a_n} \left(\sup_{g \in \mathcal{G}} \mathbb{E}_{\mathbb{P}_{a_n}}\tilde{l}_{f_{g,\gamma}}\left(\mathbf{\tilde{Z}}^{(1)}\right) - \frac{1}{b_n} \sum_{j=1}^{b_n}\tilde{l}_{f_{g,\gamma}}\left(\mathbf{\tilde{Z}}^{(j)}\right) > a_n \epsilon \right) \\ &+  2b_n \beta (a_n).
\end{align*}
%On the other hand, based on the symmetrization lemma of Vapnik and Chervonenkis,
%\begin{align*}
%\mathbb{\tilde{Q}} \left(\sup_{f \in \mathcal{F}}\mathbb{E}_{\mathbb{P}_{a_n}}\left[\tilde{l}_{f}(\mathbf{\tilde{Z}}^{(1)})\right] - \frac{1}{b_n} \sum_{j=1}^{b_n}\tilde{l}_{f}(\mathbf{\tilde{Z}}^{(j)}) > a_n \epsilon \right) \leqslant 2 \mathbb{\tilde{Q}} \left(\sup_{f \in \mathcal{F}}\frac{1}{b_n} \sum_{j=1}^{b_n} \left( \tilde{l}_{f}(\mathbf{\tilde{Z}}^{'(j)}) - \tilde{l}_{f}(\mathbf{\tilde{Z}}^{(j)}) \right) > a_n \epsilon \right),
%\end{align*}
%where $\mathbf{\tilde{Z}}^{'(j)}$ is an independent copy of $\mathbf{\tilde{Z}}^{(j)}$ for all $j$.
%Let $\bar{\mathcal{F}}$ be a subset of $\mathcal{F}$, such that $\tilde{L}_{\bar{\mathcal{F}}}$ is an $\frac{\epsilon a_n}{8}$-net of $\tilde{L}_{\mathcal{F}}$ of minimal cardinality $\mathcal{N}\left(\frac{\epsilon a_n}{8},\tilde{L}_{\mathcal{F}},d_{1,b_n}\right)$. 
%By Hoeffding's inequality,
%\begin{align*}
%P_{\boldsymbol{\sigma}_{b_n}} \left(\max_{\bar{f} \in \bar{\mathcal{F}}}\frac{1}{b_n} \sum_{j=1}^{b_n} \sigma_j \left(\tilde{l}_{\bar{f}}\left(\mathbf{\tilde{z}}^{'(j)}\right) - \tilde{l}_{\bar{f}}\left(\mathbf{\tilde{z}}^{(j)}\right) \right) > a_n \epsilon \right) \leqslant \exp\left( -\frac{b_n  \epsilon^2}{32}\right).
%\end{align*}
For any $f_1,f_2 \in \mathcal{F}$ and for any $\mathbf{z}_{a_n b_n}=(\mathbf{z}^{(j)})_{1 \leqslant j \leqslant b_n}$, define 
$$ d_{1, \mathbf{z}_{a_n b_n}}(f_1,f_2) = \frac{1}{b_n} \sum_{j=1}^{b_n} \left(\tilde{l}_{f_1}\left(\mathbf{z}^{(j)}\right)-\tilde{l}_{f_2}\left(\mathbf{z}^{(j)}\right) \right).$$ 
Now we focus on the probability in the right-hand side of the above inequality and proceed as in the proof of Theorem~\ref{theo:uniform-convergence-L1} to obtain
\begin{align*}
\mathbb{P}^{b_n}_{a_n} \left(\sup_{g \in \mathcal{G}}\mathbb{E}_{\mathbb{P}_{a_n}}\tilde{l}_{f_{g,\gamma}}\left(\mathbf{\tilde{Z}}^{(1)}\right) - \frac{1}{b_n} \sum_{j=1}^{b_n}\tilde{l}_{f_{g,\gamma}}\left(\mathbf{\tilde{Z}}^{(j)}\right) > a_n \epsilon \right) \leqslant \exp\left( -\frac{b_n  \epsilon^2}{32}\right) \mathcal{N}_1\left(\frac{\epsilon a_n}{8},\tilde{L}, 2b_n\right).
\end{align*}
According to Lemma~5.1 of \cite{Meir2000} and the Lipschitz property of $\phi_\gamma$,
\begin{align*}
\mathcal{N}_1\left(\epsilon,\tilde{L}, b_n\right) \leqslant \mathcal{N}_1\left(\frac{\epsilon \gamma}{2a_n},\mathcal{F}_{\mathcal{G},\gamma}, n\right),
\end{align*}
and the claimed result follows.
\end{proof}

\subsection*{Effective sample complexity}

To compute the effective sample complexity (i.e., an estimate for the number $2b_n=\frac{n}{a_n}$ of blocks), we need to balance the two terms in \eqref{eq:uniform-convergence-L1}: we assume that $2b_n \beta(a_n)=K_1\exp(-K_2b_n)$ 
%either to i) $\mu^{-r}_n$ for some $r \in \mathbb{N} \setminus \{ 0,1\}$; or ii) 
for some positive constants $K_1, K_2$. This imposes a constraint on $a_n$ depending on the behavior of the mixing coefficient. However, the value of the mixing coefficient is known for very few processes such as the %autoregressive moving averages model and the 
first order Markov process, for which $\beta(a_n) \leqslant \rho^{a_n}$ with $\rho \in [0,1)$ \cite{Mcdon15}. %Assume $\rho >0$. Setting $\rho^{a_n}=\mu^{-r-1}_n$, we get $a_n=\lceil \frac{(r+1) \ln b_n}{\ln (1/\rho)} \rceil$. 
In general, the mixing rate of data generating process is not known (\cite{Mcdon15} provides a way to estimate this coefficient from the data). Thus, it is usually assumed that the process is either algebraically mixing, $\beta(a_n)=\beta_0 a_n^{-k}$, or exponentially mixing, $\beta(a_n)=\beta'_0 \exp(-\beta a_n^{k'})$, for some positive $\beta_0, \beta'_0, \beta,k, k'$. Applying these assumptions, and taking benefit from Corollary~\ref{cor:new-bound} since it does not depend on the sample size, we obtain the following result which is similar to Theorem~\ref{theo:sample-complexity-iid}. The proof, being straightforward, is omitted.
\begin{theorem} Fix $\delta \in (0,1)$ and let $F=d_{\mathcal{G}_0}\left(\frac{\epsilon \gamma}{6144}\right)$. Let $K_3$ be a positive constant. For algebraically (exponentially) mixing processes, under the assumption that $\displaystyle{a_n=\left(\beta_0 \exp(K_2 b_n)/K_1\right)^{\frac{1}{k}}}$ ($\displaystyle{a_n=\left( \frac{ K_2 b_n + \ln (\beta'_0/K_1)}{\beta}\right)^{\frac{1}{k'}}}$), where
$$
b_n \geqslant \frac{1}{\min\{\epsilon^2/32,K_2\}}\left(C K_3 F \ln^2\left(\frac{C M^2}{\epsilon^2 \gamma^2} F \right)\ln\left(\frac{1}{\epsilon}\right) + \ln{\frac{4+K_1}{\delta}}\right),
$$
the probability \eqref{eq:lala} is at most $\delta$.
\end{theorem}

\section{Fat-shattering dimension and Rademacher complexity} \label{app:fat-rad}
Based on the argument of \cite{Men02}, we can derive the following result.
\begin{lemma} \label{lemma:mendelson-fat-shattering}
Let $\mathcal{F}$ be a class of functions from $\mathcal{T}$ to $\left[-M_{\mathcal{F}}, M_{\mathcal{F}}\right]$ with $M_{\mathcal{F}} \in \mathbb{R}_{+}$. For $\epsilon \in (0, M_\mathcal{F}]$, let $d\left( \epsilon \right)=\epsilon\mbox{-dim}\left(\mathcal{F}\right)$. Let $\mathbf{t}_n=(t_i)_{1 \leqslant i \leqslant n} \in \mathcal{T}^n$. For all $\epsilon \in (0, M_\mathcal{F}]$, if $\sup_{\mathbf{t}_n \in \mathcal{T}^n} \hat{R}_{n}\left(\mathcal{F}\right) \leqslant \epsilon$ for some $n \in \mathbb{N}^{*}$, then $d\left( \epsilon \right) \leqslant n$.
\end{lemma}
\begin{proof}
Let $S=\{t_i : 1 \leqslant i \leqslant d\} \in \mathcal{T}$ be the set of maximal cardinality $d$ $\epsilon$-shattered by $\mathcal{F}$. By definition of $\epsilon$-shattering, for any $\mathbf{s}_d=(s_i)_{1 \leqslant i \leqslant d} \in \{-1,1\}^d$, there exists $f_{\mathbf{s}_d}$ in $\mathcal{F}$ such that
\begin{align*}
\sum_{i=1}^{d} s_i \left(f_{\mathbf{s}_d}(t_i) - u(t_i)\right) \geqslant d\epsilon.
\end{align*}
It implies
\begin{align*}
\forall \mathbf{s}_d \in \{-1,1\}^d, \quad \sup_{f \in \mathcal{F}} \sum_{i=1}^{d} s_i \left(f(t_i) - u(t_i)\right) \geqslant d\epsilon.
\end{align*}
Then,
\begin{align*}
\frac{1}{2^d}\sum_{\mathbf{s}_n \in \{-1,1\}^d} \sup_{f \in \mathcal{F}} \sum_{i=1}^{d} s_i \left( f(t_i) - u(t_i) \right) \geqslant d\epsilon,
\end{align*}
which is equivalent to
\begin{align*}
\mathbb{E}_{\boldsymbol{\sigma}_d} \sup_{f \in \mathcal{F}} \sum_{i=1}^{d} \sigma_i \left( f(t_i) - u(t_i) \right) \geqslant d\epsilon.
\end{align*}
Since Rademacher variables are centered, the above bound reduces to
\begin{align*}
\frac{1}{d}\mathbb{E}_{\boldsymbol{\sigma}_d}\sup_{f \in \mathcal{F}} \sum_{i=1}^{d} \sigma_i f(t_i) \geqslant \epsilon.
\end{align*}
It follows that if for some $n \in \mathbb{N}^{*}$,
\begin{align*}
\frac{1}{n}\sup_{\mathbf{t}_n \in \mathcal{T}^n}\mathbb{E}_{\boldsymbol{\sigma}_n}\sup_{f \in \mathcal{F}} \sum_{i=1}^{n} \sigma_i f(t_i) \leqslant \epsilon,
\end{align*}
then 
$d \leqslant n$.
\end{proof}

\pagebreak

% that's all folks
\end{document}